\documentclass[runningheads]{llncs}
\usepackage[T1]{fontenc}
\usepackage{amssymb, amsmath, bm, latexsym, comment}
\usepackage{xcolor}
\usepackage{graphicx}
\usepackage{setspace}
\usepackage{verbatim}
\usepackage{array}
\usepackage{cite}
\usepackage{booktabs}
\usepackage{hyperref}
\usepackage[capitalize]{cleveref}
\usepackage{arydshln}
\usepackage{multirow}
\usepackage{dsfont}
\usepackage{mathtools}
\usepackage{multirow}
\usepackage{hyphenat}
\usepackage{diagbox}
\usepackage{tikz}
\usepackage{enumitem}

\graphicspath{{./figures/}}

\newcolumntype{L}[1]{>{\raggedright\let\newline\\\arraybackslash\hspace{0pt}}m{#1}}
\newcolumntype{C}[1]{>{\centering\let\newline\\\arraybackslash\hspace{0pt}}m{#1}}
\newcolumntype{R}[1]{>{\raggedleft\let\newline\\\arraybackslash\hspace{0pt}}m{#1}}

\DeclarePairedDelimiterX{\infdivx}[2]{\big[}{\big]}{%
  #1\;\delimsize\|\;#2%
}
\newcommand{\KLdiv}{D_{\text{KL}}\infdivx}
\def\mid{\,|\,}
\newcommand{\circsmall}{\,\raisebox{1pt}{\tikz \draw[line width=0.6pt] circle(1.2pt);}\,}

\DeclarePairedDelimiter\abs{\lvert}{\rvert}%
\DeclarePairedDelimiter\norm{\lVert}{\rVert}%

\makeatletter
\let\oldabs\abs
\def\abs{\@ifstar{\oldabs}{\oldabs*}}
\let\oldnorm\norm
\def\norm{\@ifstar{\oldnorm}{\oldnorm*}}
\makeatother

\begin{document}

\title{BInGo: Bayesian Intrinsic Groupwise Registration via Explicit Hierarchical Disentanglement
\thanks{X. Wang and X. Luo contributed equally to this work.}}

\titlerunning{BInGo: Bayesian Intrinsic Groupwise Registration}

\author{Xin Wang\inst{1,2}
\and Xinzhe Luo\inst{1}\and Xiahai Zhuang\inst{1}}

\authorrunning{X. Wang, X. Luo, and X. Zhuang}   

\tocauthor{}

\institute{School of Data Science, Fudan University, Shanghai, China \and
Department of Electrical and Computer Engineering, University of Washington, Seattle, United States}

\maketitle

\begin{abstract}
Multimodal groupwise registration aligns internal structures in a group of medical images. 
Current approaches to this problem involve developing similarity measures over the joint intensity profile of all images, which may be computationally prohibitive for large image groups and unstable under various conditions.
To tackle these issues, we propose BInGo, a general \emph{unsupervised} hierarchical Bayesian framework based on deep learning, to learn \emph{intrinsic} structural representations to measure the similarity of multimodal images. 
Particularly, a variational auto-encoder with a novel posterior is proposed, which facilitates the \emph{disentanglement} learning of structural representations and spatial transformations, and characterizes the imaging process from the common structure with shape transition and appearance variation.
Notably, BInGo is scalable to learn from small groups, whereas being tested for large-scale groupwise registration, thus significantly reducing computational costs.
We compared BInGo with five iterative or deep-learning methods on three public intrasubject and intersubject datasets, i.e. BraTS, MS-CMR of the heart, and Learn2Reg abdomen MR-CT, 
and demonstrated its superior accuracy and computational efficiency, even for very large group sizes (e.g., over 1300 2D images from MS-CMR in each group).
\end{abstract}

\section{Introduction}
Multimodal groupwise registration aims to align multimodal images into a common structural space. 
Unlike conventional pairwise registration which aligns moving images separately to a fixed image, groupwise registration can ameliorate the bias from designating a reference image, by estimating a common space to which all the images are co-registered. 
Therefore, it has become an essential task in multivariate image analysis, including longitudinal research, atlas construction, motion estimation, and population studies \cite{journal/neuroimage/liao2012,journal/neuroimage/joshi2004,journal/media/metz2011,journal/neuroimage/geng2009}.

Conventional iterative methods usually estimate the desired spatial transformations by optimizing intensity-based similarity measures  \cite{journal/tpami/learned2005,journal/tip/orchard2009,journal/tpami/wachinger2012,journal/media/polfliet2018,journal/tpami/luo2022}.
Recently, unsupervised deep-learning methods attempt to realize groupwise registration in an end-to-end fashion \cite{conference/ipmi/che2019,conference/icip/he2020}.
For instance in \cite{conference/ipmi/che2019}, the authors devised a network to optimize the conditional template entropy (CTE) introduced in \cite{journal/media/polfliet2018}.
These learning-based models estimate conventional similarity measures developed for iterative registration using stochastic gradient methods, 
and hence may inherit the same suboptimal performance due to insufficient modeling of the image similarity. 
Besides, these measures rely on correctly characterizing the joint intensity profile over the entire image group, which may be computationally prohibitive and applicability-limited due to large group sizes.


In this work, however, we shift attention from devising similarity metrics to learning the underlying generative imaging process.
Specifically, we establish a probabilistic generative model for the observed images, in which transformations and the common structure are disentangled as latent variables. 
In this way, groupwise registration can be achieved by unsupervised variational auto-encoding:
the encoder extracts structural representations of images and then infers the spatial deformations; the decoder imitates the imaging process by reconstructing original images from estimated common structure and the inverse deformations.
The contributions of this work can be summarized as follows:
\begin{enumerate}[label=(\arabic*)]
  \item We propose BInGo, a theoretically-grounded unsupervised Bayesian framework for groupwise registration, which learns intrinsic structural similarity of multimodal images by a principled variational distribution, and is capable of disentangling structural representations from image appearance.
  \item {BInGo is scalable to be trained with small image groups while being applied to large-scale and variable-size test groups, significantly improving applicability and computational efficiency.}
  \item We demonstrate the superiority of BInGo over similarity-based methods on three multimodal intrasubject and intersubject datasets.
\end{enumerate}
To the best of our knowledge, this is the first work that realizes scalable multimodal groupwise registration by disentangled representation learning.

\section{Methodology}
Let the original image group be $\bm{X}=(X_m)_{m=1}^M$, with $M$ the number of modalities and $X_m:\mathbb{R}^d\supset\mathrm{\Omega}_m\rightarrow\mathbb{R}$. 
Groupwise registration aims to find spatial transformations $\bm{\phi}=(\phi_m:\mathbb{R}^d\supset\mathrm{\Omega}\rightarrow\mathrm{\Omega}_m)_{m=1}^M$, such that the aligned images $(X_m\circsmall\phi_m)_{m=1}^M$ share a \emph{common} structural representation $\bm{Z}$. 

We assume $X_m$ is generated from $\bm{Z}$ and $\phi_m^{-1}$ by a \emph{transformation\hyp{}equivariant} imaging process $f_m$ that contains modality\hyp{}specific appearance information, i.e. 
\begin{equation}\label{eq:disentangle}
  X_m = f_m(\bm{Z}\circsmall\phi_m^{-1})=f_m(\bm{Z})\circsmall\phi_m^{-1}.
\end{equation}
Therefore, a 3-step unsupervised auto-encoding scheme can be formed: 
I) By modeling $f_m^{-1}$ we could encode \emph{individual} structural representation for $X_m$ as $\bm{Z}\circsmall\phi_m^{-1}=f_m^{-1}(X_m)$, from which $\bm{\phi}$ could be estimated more easily. 
II) $\bm{Z}$ could be then inferred from $\bm{Z}=f_m^{-1}(X_m\circsmall\phi_m)$ following the equivariance assumption. 
III) By modeling $f_m$ we could reconstruct $X_m$ from the estimated $\bm{Z}$ and $\phi_m^{-1}$. 

The following establish BInGo to realize this scheme by variational and disentangled auto-encoding, thus learning $\bm{\phi}$ in a unified and end-to-end fashion.

\subsection{Hierarchical Bayesian Inference}\label{sec:HGM}

\begin{figure}[t]
    \centering
    \includegraphics[height=2.8cm]{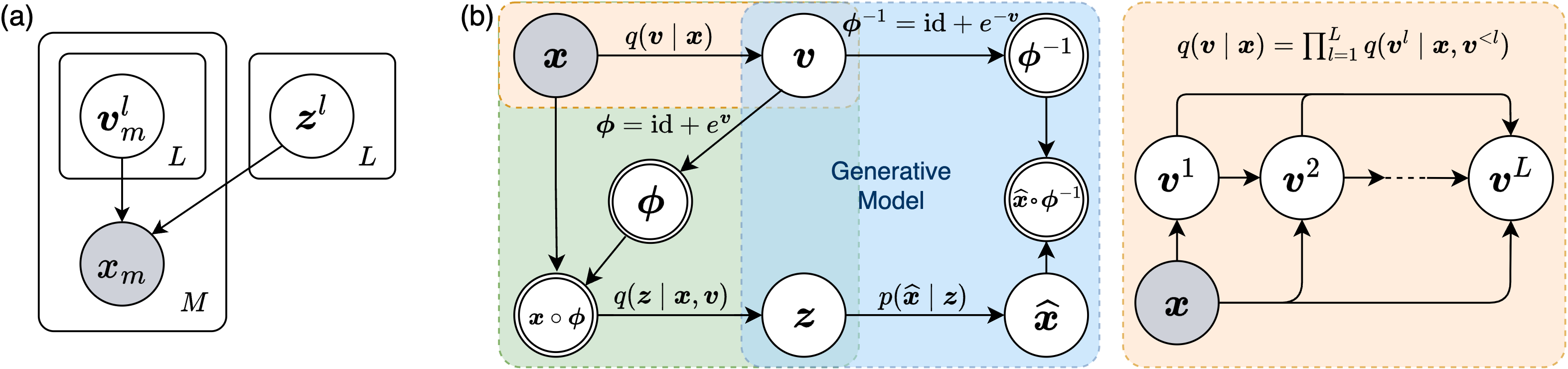}
  \caption{The proposed Bayesian framework. \textbf{(a)} Graphical model of the imaging process. \textbf{(b)} Inference model for the velocity fields $\bm{v}$ (orange), the common structural representation $\bm{z}$ (green), and the reconstruction (blue). Random variables are in circles, deterministic variables are in double circles, and observed variables are shaded.}
  \label{fig:graph}
\end{figure}

We first formalize the estimation of spatial deformations $\bm{\phi}=(\phi_m)_{m=1}^M$ through Bayesian inference, as illustrated in \cref{fig:graph}.
Specifically, let $\bm{x}=(x_m)_{m=1}^M$ be a sample of $\bm{X}$.
We decompose the latent variables generating $\bm{x}$ into two \emph{independent} subgroups:
1) the common structural representation $\bm{z}$, and
2) the stationary velocity fields $\bm{v}=(\bm{v}_m)_{m=1}^M$ that parameterize $\bm{\phi}$ \cite{journal/NI/ashburner2007}.
Thus, following the variational Bayes framework, the objective function to maximize is the evidence lower bound (ELBO) of the log-likelihood, which takes the form
\begin{equation}\label{eq:ELBO}
  \begin{aligned}
    \mathcal{L}(\bm{x}) \triangleq&\  \mathbb{E}_{q(\bm{z},\bm{v}\mid \bm{x})}\big[\log p(\bm{x}\mid\bm{z},\bm{v})\big] - \KLdiv{q(\bm{z},\bm{v}\mid \bm{x})}{p(\bm{z})\,p(\bm{v})}
  \end{aligned}
\end{equation}
where $q(\bm{z},\bm{v}\mid\bm{x})=q(\bm{v}\mid\bm{x})\,q(\bm{z}\mid \bm{v}, \bm{x})$ defines the variational posterior distribution, and $D_{\text{KL}}$ is the Kullback-Leibler (KL) divergence.
After optimizing the ELBO, the desired velocity fields $\widehat{\bm{v}}$ can be inferred via maximum a posteriori.

Furthermore, we express the latent variables with $L$ hierarchical levels \cite{conference/nips/vahdat2020}, i.e. $\bm{z}=(\bm{z}^l)_{l=1}^L$ and $\bm{v}_{m}=(\bm{v}_m^l)_{l=1}^L$, and a higher level indicates a finer-scale (larger) resolution. 
Thus, the total deformation $\phi_m=\phi_m^1\circsmall\cdots\circsmall\phi_m^L$ can be deterministically computed by exponentiation of velocity fields ${\phi}_m^l=\mathrm{id}+\exp({\bm{v}}_m^l)$, where $\frac{\partial}{\partial t}\phi_m^l(\bm{\omega}, t)=\bm{v}_m^l(\phi_m^l(\bm{\omega}, t))$, $\forall\,\bm{\omega}\in\mathrm{\Omega}$. 
The hierarchical strategy allows to model a complex deformation by several simpler and easier-to-learn ones.

To simplify the KL, we introduce additional independence assumptions: 
1) both the prior and posterior of the velocities factorize among different images, i.e. $p(\bm{v}^l\mid\bm{v}^{<l})=p(\bm{v}^l)=\prod_{m=1}^M p(\bm{v}_m^l)$ and $q(\bm{v}^l\mid\bm{x},\bm{v}^{<l})=\prod_{m=1}^M q(\bm{v}_m^l\mid\bm{x},\bm{v}^{<l})$, where $<l$ denotes the group of latent variables in levels less than $l$, and
2) the common structure $\bm{z}^l$ can be inferred directly from the observed images and the estimated velocity fields, i.e. $q(\bm{z}^l\mid\bm{x},\bm{v},\bm{z}^{<l})=q(\bm{z}^l\mid\bm{x},\bm{v})$, since $\{\bm{x},\bm{v}\}$ can determine registered images, thus containing all information about $\bm{z}^l$ for any $l$.

Taking into account these assumptions, the KL divergence can be written as
\begin{equation}\label{eq:KL}
  \begin{aligned}
    &\KLdiv{q(\bm{v}\mid\bm{x})\,q(\bm{z}\mid \bm{v}, \bm{x})}{p(\bm{v})\,p(\bm{z})} \\
    =&\sum_{m=1}^M\left[\sum_{l=1}^L\mathbb{E}_{q(\bm{v}_m^{<l}\mid\bm{x})}\Big\{\KLdiv{q(\bm{v}_m^l\mid\bm{x},\bm{v}^{<l})}{p(\bm{v}_m^l)}\Big\}\right]\qquad\qquad \text{(i)}\ \\
    &+ \mathbb{E}_{q(\bm{v}\mid\bm{x})}\left[\sum_{l=1}^L\mathbb{E}_{q(\bm{z}^{<l}\mid\bm{x},\bm{v})}\Big\{\KLdiv{q(\bm{z}^l\mid\bm{x},\bm{v})}{p(\bm{z}^l\mid\bm{z}^{<l})}\Big\}\right]\ \text{(ii)},
  \end{aligned}
\end{equation}
where we prescribe $p(\bm{z}^1\mid\bm{z}^{<1})\triangleq p(\bm{z}^1)$ and $q(\bm{v}_m^{<1}\mid\bm{x})= q(\bm{z}^{<1}\mid\bm{x},\bm{v})\triangleq 1$ for simplicity.
The key idea here is: the overall KL divergence is decomposed \emph{w.r.t.} (i) the velocity fields $\bm{v}$, and (ii) the common structure $\bm{z}$.
The former serves as regularization to ensure diffeomorphism, fulfilled by the constraint introduced in \cite{journal/media/dalca2019};
the latter is intended to estimate the structural similarity among the images, as is detailed in the next subsection.

\subsection{Intrinsic Similarity over Structural Representations}
For registration, it is crucial to efficiently measure the structural similarity, which is achieved in this work by learning instead of a pre-defined metric. To this end, we propose to learn multilevel ``expert'' distributions $q_m(\bm{z}^l\mid x_m,\bm{v}_m)$, which serve as $f_m^{-1}$ (i.e., the inverse of imaging processes) to extract the structural representations of warped images. We further assume $q(\bm{z}^l\mid\bm{x},\bm{v})$ and $p(\bm{z}^l\mid\bm{z}^{<l})$ to be the \emph{geometric} mean \cite{journal/media/lorenzen2006} and \emph{arithmetic} mean \cite{conference/nips/shi2019} of the experts, respectively:
\begin{equation}\label{eq:joint_post}
  q(\bm{z}^l|\,\bm{x},\bm{v})\propto \left[\prod_{m=1}^M q_m(\bm{z}^l|\, x_m,\bm{v}_m)\right]^{\frac{1}{M}},\ 
  p(\bm{z}^l|\bm{z}^{<l})\triangleq \frac{1}{M}\sum_{m=1}^M q_m(\bm{z}^l|\, x_m,\bm{v}_m).
\end{equation}
Therefore, the KL in \cref{eq:KL}(ii) essentially measures the \emph{intrinsic} (dis)similarity, i.e., the dissimilarity of the experts (intrinsic structural representations), whose minimization, as a part of the maximization of the ELBO, encourages the experts to be identical, thus forcing the multilevel posteriors $q(\bm{z}^l\mid\bm{x},\bm{v})$ to represent the common structure. 
Meanwhile, the velocity could be learned in tandem.

We model the experts $q_m(\bm{z}^l\mid x_m,\bm{v}_m)$ as Gaussians $\mathcal{N}(\bm{\mu}_m^l,\bm{\Sigma}_m^l)$, and thus so is the joint posterior, i.e. $q(\bm{z}^l\mid\bm{x},\bm{v})=\mathcal{N}(\bm{\mu}^l,\bm{\Sigma}^l)$ with
\begin{equation}\label{eq:geometric_mean}
  \bm{\Sigma}^l = M\cdot\left[\sum_{m=1}^M\left(\bm{\Sigma}_m^l\right)^{-1}\right]^{-1},\quad
  \bm{\mu}^l = \frac{\bm{\Sigma}^l}{M}\cdot\sum_{m=1}^M \bm{\mu}_m^l\left(\bm{\Sigma}_m^l\right)^{-1}.
\end{equation}
In light of the computational intractability of the KL divergence involving Gaussian mixture distributions, we further exploit its convexity to obtain
\begin{equation}\label{eq:kl_convexity}
  \KLdiv{q(\bm{z}^l\mid\bm{x},\bm{v})}{p(\bm{z}^l\mid\bm{z}^{<l})}\leqslant \frac{1}{M}\sum_{m=1}^M \KLdiv{q(\bm{z}^l|\,\bm{x},\bm{v})}{q_m(\bm{z}^l|\, x_m,\bm{v}_m)}.
\end{equation}
Hence, the minimization of the KL in \cref{eq:KL}(ii) is actually approximated by minimizing the right-hand side of \cref{eq:kl_convexity}, which has a closed-form expression.

\subsection{Explicit Disentanglement with Neural Networks}\label{sec:NN}
\begin{figure}[t]
  \centering
  \includegraphics[width=\textwidth]{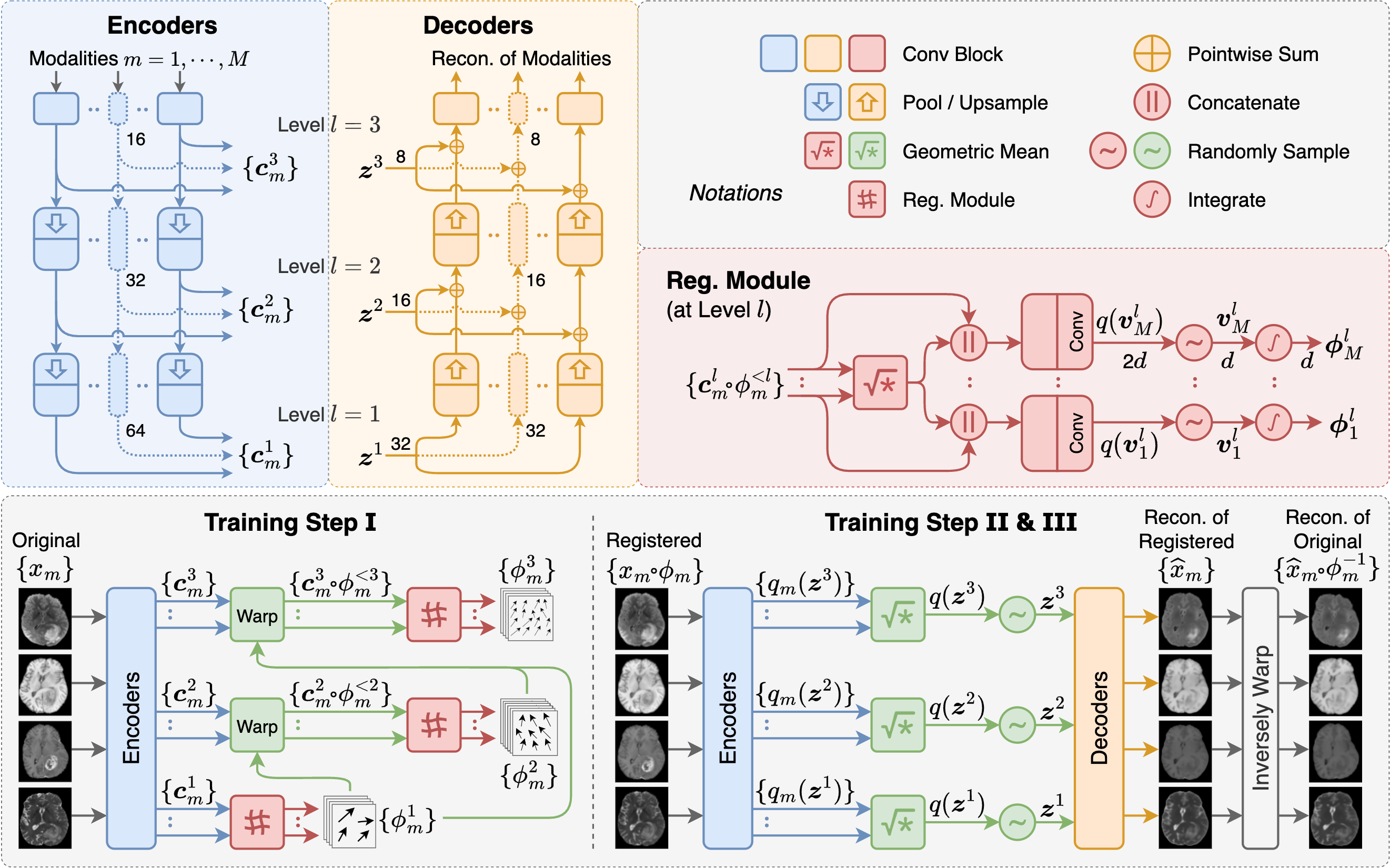}
  \caption{Architecture ($L=3$) of BInGo for explicit disentanglement. The channel numbers are indicated around particular feature maps (arrows).}
  \label{fig:network}
\end{figure}

The ELBO has been decomposed into three terms: reconstruction of the original images (the first term in \cref{eq:ELBO}), the KL divergence for velocity fields to ensure diffeomorphism (\cref{eq:KL}(i)), and the KL divergence for the intrinsic (dis)similarity to optimize the registration (\cref{eq:KL}(ii)), with weights (hyperparameters) $(\lambda_i)_{i=1}^3$. To estimate the ELBO, we propose a dedicated hierarchical variational auto-encoder (VAE) as the inference model in \cref{fig:graph}(b), whose architecture with the number of hierarchy $L=3$ is depicted in \cref{fig:network}. 
The network learns $p(\bm{x} \mid \bm{z}, \bm{v}), q\left(\bm{v}_{m}^{l} \mid \bm{x}, \bm{v}^{<l}\right)$ and $q_{m}\left(\bm{z}^{l} \mid x_{m}, \bm{v}_{m}\right)$, and the expectations in the ELBO are estimated by Monte Carlo sampling, similar to traditional VAEs.

The motif of BInGo is to explicitly disentangle the common structure, spatial transformations and appearance information from multimodal images, thus realizing \cref{eq:disentangle}.
To this end, the network comprises three types of cooperative modules: 
1) $M$ encoders that extract multi-level \emph{transformation-equivariant} structural representations $\bm{c}_m^l$, corresponding to the inverse imaging function $f_m^{-1}$,
2) multi-level registration (Reg) modules that infer spatial transformations from individual structural representations, corresponding to estimating $\bm{\phi}$ from $\{\bm{z}\circsmall\phi_m^{-1}\}_{m=1}^M$,
and 
3) $M$ decoders, with modality-specific appearance information embedded as parameters, which reconstruct registered images from learned common structure, corresponding to the imaging function $f_m(\bm{z})$.

Particularly, the Reg module at level $l$ assumes the coarser-scale deformations $\phi_m^{<l}\triangleq\phi_m^{1}\circsmall\cdots\circsmall\phi_m^{l-1}$ have been resolved, and thus takes the individual representations $\bm{c}_m^l(x_m)\circsmall\phi_m^{<l}$ as inputs, which are like ``partially distorted'' version of the expert distributions, since the experts are representations of fully registered images $x_m\circsmall\phi_m$. Thus, in the same spirit of \cref{eq:joint_post}, the Reg module computes their geometric mean as the ``partially common'' structure, and then concatenates the mean and $\bm{c}_m^l(x_m)\circsmall\phi_m^{<l}$ to infer $q(\bm{v}_{m}^{l}\mid \bm{x},\bm{v}^{<l})$ for each $m$ separately.

Therefore, BInGo can realize the aforementioned 3-step unsupervised scheme induced by \cref{eq:disentangle}, as follows:

\noindent\textbf{I. Bottom-up Inference of Velocity Fields.}
The encoders first produce multilevel feature maps $\bm{c}_m^l(x_m)$ as the \emph{individual} structural representations of original images.
Then, up from the bottom ($l$ increasing from 1 to $L$), given that $\phi_m^{<l}$ have been inferred ($\phi_m^{<1}\triangleq \mathrm{id}$), the feature maps are warped to obtain $\bm{c}_m^l(x_m)\circsmall\phi_m^{<l}$. Then, the Reg module at level $l$ can infer $q(\bm{v}_{m}^{l}\mid \bm{x},\bm{v}^{<l})$, from which $\bm{v}_m^l$ is sampled and $\phi_m^l$ is finally computed through integration.

\noindent\textbf{II. Top-down Inference of Common Structures.}
As the total deformations $\phi_m=\phi_m^1\circsmall\cdots\circsmall\phi_m^L$ have been inferred, the encoders are fed \emph{again} with warped images $\bm{x}\circsmall\bm{\phi}\triangleq(x_m\circsmall\phi_m)_{m=1}^M$ to produce the experts $q_{m}(\boldsymbol{z}^{l} \mid x_{m}, \boldsymbol{v}_{m})=\bm{c}_m^l(x_m\circ \phi_m)$ in a top-down ($l$ decreasing from $L$ to $1$) manner. The posterior $q\left(\boldsymbol{z}^{l} \mid \boldsymbol{x}, \boldsymbol{v}\right)$ is then computed by \cref{eq:joint_post}, from which the modality-invariant common structural representation $\boldsymbol{z}=(\boldsymbol{z}^{l})_{l=1}^L$ is sampled. Note that we can avoid using encoders twice by warping $\bm{c}_m^l(x_m)$ to compute the experts as $\bm{c}_m^l(x_m)\circ \phi_m$ (thanks to equivariance), but this requires more computational cost. 

\noindent\textbf{III. Disentangled Auto-Encoding.}
Based on the common structure $\bm{z}$ inferred by encoders, the decoders reconstruct the registered images $\widehat{\bm{x}}=(\widehat{x}_m)_{m=1}^M$. Then reconstructed original images $\widehat{\bm{x}}\circsmall\bm{\phi}^{-1}\triangleq(\widehat{x}_m\circsmall\phi_m^{-1})_{m=1}^M$ are obtained by inverse deformations. We utilize (negative) L1 loss between the original images and their reconstructions to estimate the first (reconstruction) term in \cref{eq:ELBO}. In addition, to better disentangle structure and appearance, the convolutional layers of the network are \emph{shared} across modalities, while \emph{domain-specific} batch normalization layers (BNs) encode appearance information for each modality \cite{conference/cvpr/chang2019}.

\subsection{Towards Large-Scale and Variable-Size Groupwise Registration}
Most conventional methods optimize similarity measures defined over the entire image group, making the computation for large groups formidable. Besides, learning-based models that rely on these measures require groups to have the same size. These drawbacks significantly reduce their applicability.

\begin{figure}[t]
  \centering
  \includegraphics[width=\textwidth]{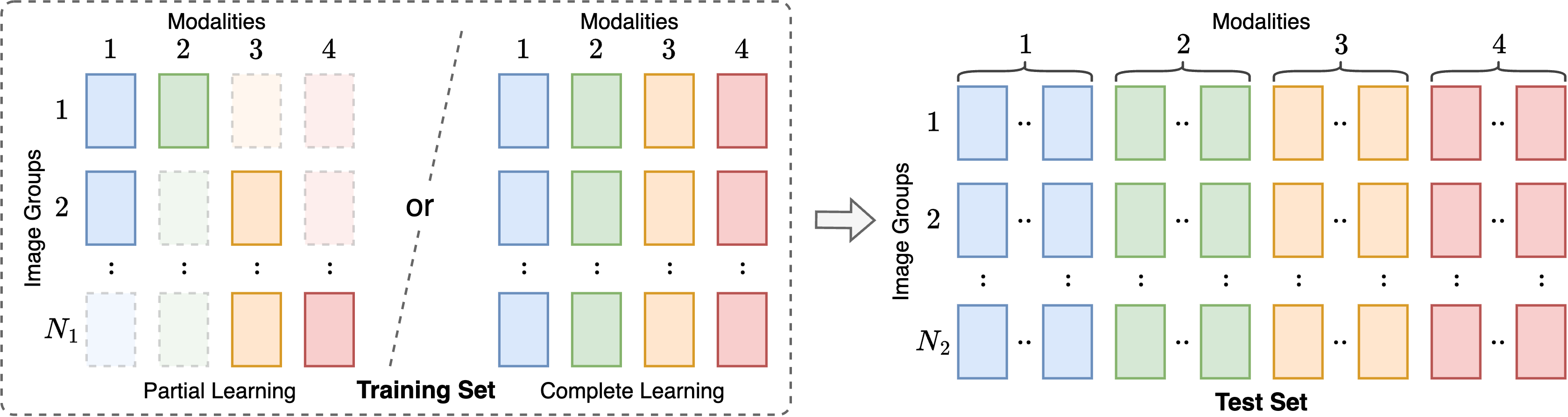}
  \caption{Partial and complete learning schemes of our model for large-scale and variable-size groupwise registration.}
  \label{fig:partial_learn}
\end{figure}

However, BInGo is scalable to handle multimodal image groups with various sizes for training and test. As shown in \cref{fig:partial_learn}, it can be trained with either bimodal image pairs or image groups of complete modalities, referred to as \emph{partial} or \emph{complete} learning, respectively. Trained in either way, BInGo can be flexibly applied to larger unseen test groups with arbitrary numbers of images, such that computational efficiency of training is substantially boosted.

The scalability of our model is conducive in two scenarios:

\noindent\textbf{Intrasubject Images with Missing Modalities.}
The goal is to co-register multimodal scans for each patient. In practice, there could be many patients with different missing modalities. BInGo can be trained with these heterogeneous datasets via partial learning, while performing groupwise registration for test subjects with complete modalities.

\noindent\textbf{Intersubject Populations.}
Intersubject groupwise registration, which is crucial in atlas construction and population analysis, could involve plenty of images for every modality. 
Conventional methods to this problem rely on iterative optimization and suffer from computational complexity scaling with image group size, while BInGo can greatly relieve training burden and be subsequently applied to test groups of complete populations in one shot. 

\section{Experiments and Results}\label{sec:experiments&results}
\subsection{Datasets and Preprocessing}
\noindent\textbf{BraTS-2021.}
The dataset provides 3D pre-operative T1, T1Gd, T2 and T2-FLAIR MR scans of patients with glioblastoma \cite{journal/tmi/menze2014,report/arxiv/baid2021}.
We randomly selected 300/50/150 patient sets for training/validation/test.
The volumes were downsampled into $2\times 2\times 2 \text{mm}^3$ with ROI of size $80\times 96\times 80$.
As the images of each patient are pre-registered, we use synthetic free-form deformations (FFDs) with different control point spacings to simulate misalignments.

\noindent\textbf{MS-CMR.}
The MS-CMRSeg challenge \cite{journal/media/zhuang2022} provides cardiac MR sequences LGE, bSSFP, and T2 from 45 patients, which exhibit complementary information of the cardiac structure and pathology.
The images were preprocessed by affine co-registration, ROI cropping, and slice selection, producing 39/15/44 slices for training/validation/test.
We simulated worse misalignments by applying additional FFDs on original images to better demonstrate model efficacy.

\noindent\textbf{Learn2Reg Abdomen MR-CT.}
This dataset \cite{dataset/CHAOSdata2019,journal/tmi/hering2021} collects 3D MR and CT volumes.
The images were resampled into $3\times 3\times 3\text{mm}^3$ with size of $128\times 107\times 128$.

\subsection{Experimental Setups}
\noindent\textbf{Compared Methods.}
Three types of unsupervised methods for multimodal groupwise registration were compared on the \emph{intrasubject} BraTS and MS-CMR datasets:
1) similarity-based iterative methods using information-theoretic metrics CTE, APE or $\mathcal{X}$-CoReg \cite{journal/media/polfliet2018,journal/tpami/wachinger2012,journal/tpami/luo2022},
2) similarity-based deep-learning models that optimize CTE or APE using an attention residual U-Net (AttResUNet) \cite{journal/media/hu2018,conference/midl/oktay2018} as the backbone,
3) BInGo, the proposed model learning intrinsic similarity through hierarchical disentanglement.
For \emph{intersubject} population groupwise registration on Learn2Reg, the models in 2) are not applicable since they can only be trained with \emph{complete} image groups, whereas BInGo can be trained \emph{partially}.
In addition, for BraTS and MS-CMR, all baseline methods used single-level velocity fields as the transformation model; for Learn2Reg, iterative baselines performed rigid, affine and FFD registration successively for optimal accuracy, due to the severe initial misalignments.

\noindent\textbf{Implementation Details.}
Training was done through the Adam optimizer \cite{DBLP:journals/corr/KingmaB14} (learning rate: $10^{-3}$; batch size: $20/1$ for MS-CMR/other datasets). 
The experiments were conducted with PyTorch \cite{software/paszke2017} on an NVIDIA $\text{RTX}^\textnormal{TM}$ 3090 GPU.

\noindent\textbf{Evaluation Metrics.}
We reported the Dice similarity coefficient (DSC) averaged over all pairwise combinations of the segmentation masks for registered images.
For the BraTS dataset, as the ground-truth misalignments were available, we also reported the groupwise warping index (gWI) \cite{journal/tpami/luo2022}, which would reduce to zero if the misaligned images were perfectly co-registered.

\subsection{Results}
\begin{table}[t]
  \scriptsize
  \centering
  \caption{\small Results on intrasubject image groups from BraTS and MS-CMR. 
  The mean values and standard deviations are presented for the gWI (in voxels) and DSC, with top 2 bolded. The number of parameters (in millions) of each model for each dataset are also reported. Statistically significant improvement ($p<0.05$ for one-sided paired $t$-tests) of BInGo with complete$^{\dagger}$ or partial$^{*}$ learning was marked with daggers or asterisks, respectively.}
  \begin{tabular}{C{2.5cm}|C{2.1cm}C{2.1cm}C{1.3cm}C{2.1cm}C{1.3cm}}
    \toprule
    \multirow{2}{*}{\small Method} & \multicolumn{3}{c}{BraTS} & \multicolumn{2}{c}{MS-CMR} \\
    \cmidrule(lr){2-4}
    \cmidrule(lr){5-6}
    & \multicolumn{1}{c}{DSC $\uparrow$} & \multicolumn{1}{c}{gWI $\downarrow$} & \multicolumn{1}{c}{\#Params.} & \multicolumn{1}{c}{DSC $\uparrow$} & \multicolumn{1}{c}{\#Params.}  \\
    \midrule
    None & $0.610\pm 0.150^{\dagger*}$ & $1.430\pm 0.644^{\dagger*}$ & --- & $0.722\pm 0.101^{\dagger*}$ & --- \\
    \hdashline\noalign{\vskip 0.5ex}
    APE \cite{journal/tpami/wachinger2012} & $\bm{0.726\pm 0.078}$ & $\bm{0.596\pm 0.149}$ & 7.373 & $0.811\pm 0.072^{\dagger*}$ & 0.154 \\ 
    CTE \cite{journal/media/polfliet2018} & $0.561\pm 0.148^{\dagger*}$ & $1.087\pm 0.411^{\dagger*}$ & 7.373 & $0.816\pm 0.077^{\dagger*}$  & 0.154 \\
    $\mathcal{X}$-CoReg \cite{journal/tpami/luo2022} & $0.707\pm 0.089^{\dagger}$ & $0.697\pm 0.212^{\dagger}$ & 7.373 & $0.840\pm 0.077^{\dagger*}$ & 0.154 \\
    \hdashline\noalign{\vskip 0.5ex}
    APE+AttResUNet & $0.693\pm0.078^{\dagger}$ & $0.757\pm0.153^{\dagger*}$ & 22.955 & $0.846\pm 0.048^{\dagger*}$ & 8.036\\
    CTE+AttResUNet & $0.659\pm0.096^{\dagger*}$ & $0.916\pm0.210^{\dagger*}$ & 22.955 & $0.874\pm0.043^{\dagger*}$ & 8.036\\
    \hdashline\noalign{\vskip 0.5ex}
    BInGo (complete$^{\dagger}$) & $\bm{0.717\pm 0.068}$ & $\bm{0.596\pm 0.132}$ & 13.429 & $\bm{0.887\pm 0.033}$ & 4.516\\
    BInGo (partial$^*$) & $0.693\pm 0.075$ & $0.709\pm 0.172$ & 13.429 & $\bm{0.877\pm0.042}$ & 4.516\\
    \bottomrule
  \end{tabular}
  \label{tab:result_intra}
\end{table}

\begin{table}[t]
  \scriptsize
  \centering
  \caption{\small Results on intersubject population groups from Learn2Reg. The means and standard deviations of DSCs versus the numbers of images in each group are reported.}
  \begin{tabular}{C{3cm}|C{2cm}C{2cm}C{2cm}C{2cm}}
    \toprule
    \backslashbox{Method}{\#Images} & 2 & 4 & 8 & 16  \\
    \midrule
    None & $0.396\pm 0.168$ & $0.386\pm0.070$  & $0.319\pm0.007$ & $0.306\pm0.000$ \\
    \hdashline\noalign{\vskip 0.5ex}
    APE \cite{journal/tpami/wachinger2012} & $0.586\pm 0.376$ & $0.574\pm 0.261$ & N/A & N/A \\
    CTE \cite{journal/media/polfliet2018}  & $0.609\pm 0.306$ & $0.088\pm 0.036$ & N/A & N/A \\
    $\mathcal{X}$-CoReg \cite{journal/tpami/luo2022} & $0.675\pm 0.329$ & $0.567\pm 0.224$ & N/A & N/A \\
    \hdashline\noalign{\vskip 0.5ex}
    BInGo (partial$^*$) & $\bm{0.781\pm0.108}$ & $\bm{0.715\pm0.122}$ & $\bm{0.677\pm0.059}$ & $\bm{0.645\pm0.000}$  \\ 
    \bottomrule
  \end{tabular}
  \label{tab:result_inter}
\end{table}

\begin{figure}[t]
  \centering
    \includegraphics[width=\textwidth]{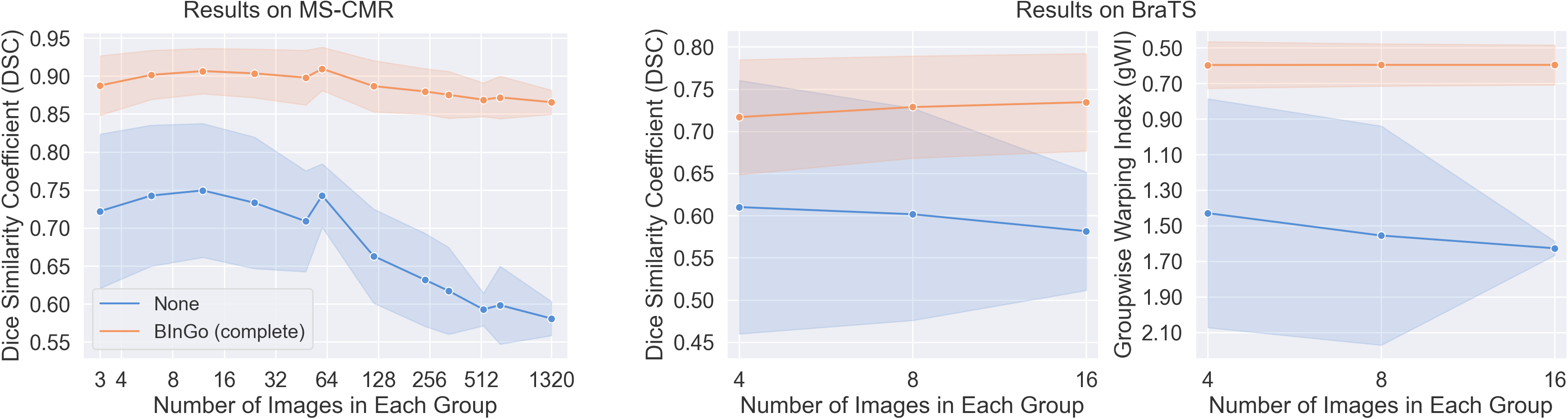}
  \caption{Results (mean values with one standard deviation bands) before (blue) and after (orange) registration by trained BInGo for images groups with different sizes.}
    \label{fig:large_group_results}
\end{figure}

\begin{figure}[h!]
  \centering
  \includegraphics[width=\textwidth]{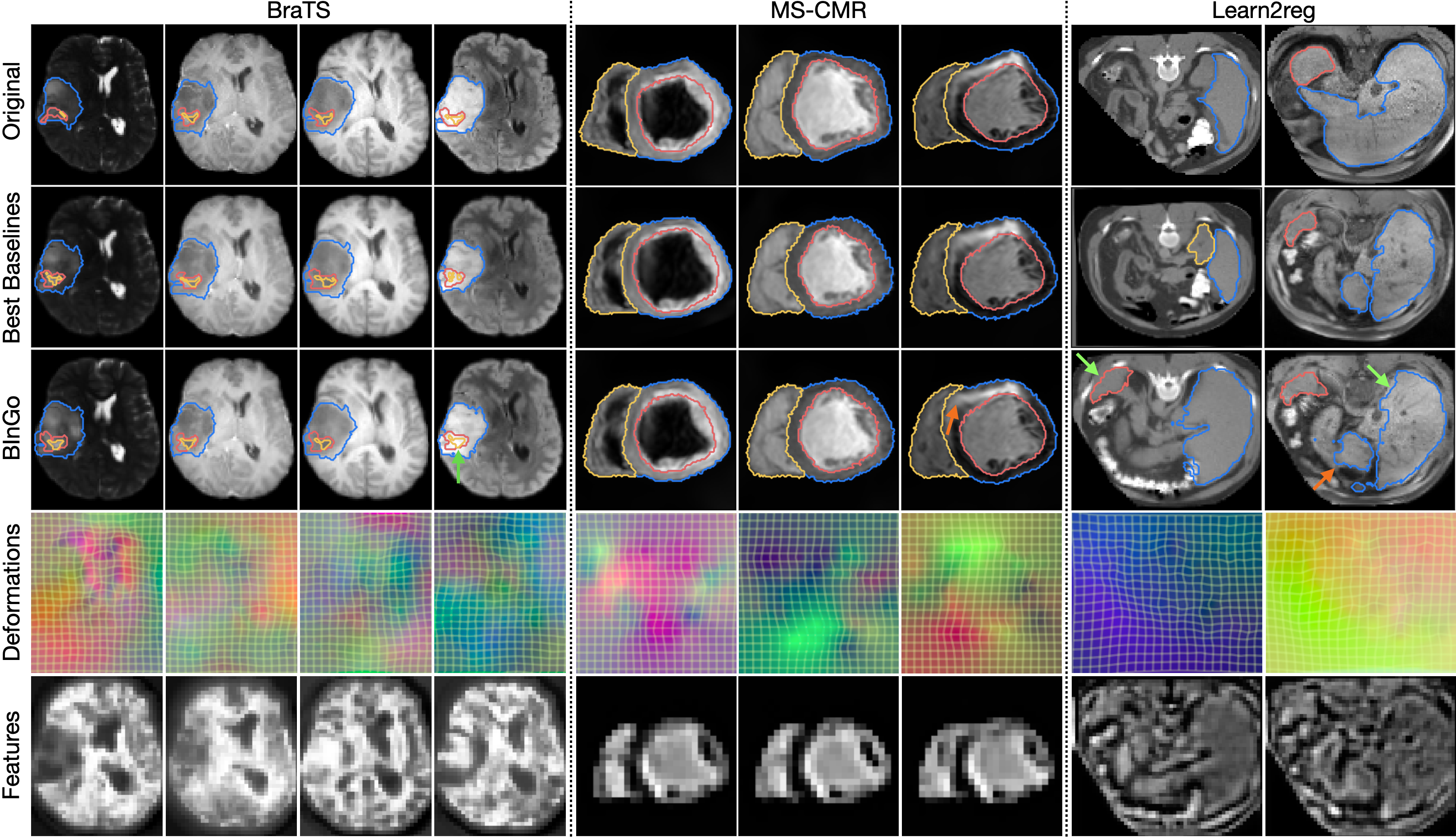}
  \caption{Example image groups (with segmentation contours overlaid) before or after registration using the best baselines or BInGo. Deformations and example features (extracted from registered images) output by BInGo are also presented. Green and orange arrows indicate some areas of significant improvement (compared to the best baselines) and relatively poor performance, respectively.}
  \label{fig:vis_results}
\end{figure}

\noindent\textbf{Intrasubject Image Groups.}
\cref{tab:result_intra} presents registration accuracy on the BraTS and MS-CMR datasets. 
BInGo (with complete$^{\dagger}$ or partial$^{*}$ learning) worked consistently better than similarity-based iterative or learning approaches, except for the APE-based iterative method on BraTS.
Notably, BInGo could achieve superior performance to deep-learning baselines with only half of training parameters, even though their backbone AttResUNet is more advanced than the vanilla UNet-like structure used in BInGo. 
Furthermore, the proposed partial learning strategy could perform better than most baselines with significantly lower computational demands, which potentiates large-scale end-to-end groupwise registration.

\noindent\textbf{Intersubject Population Image Groups.}
For Learn2Reg, we merged every $K$ test MR-CT pairs to form larger groups (each containing $M'=2K$ images to co-register) as test sets of intersubject populations. \cref{tab:result_inter} presents the results versus different $M'$. All iterative approaches failed when the groups became large due to excessive GPU cost, while our partial learning strategy worked consistently better and maintained a good performance for large groups. Note that the deep-learning baselines are not capable of handling test groups with varying sizes, and thus not presented. 

\noindent\textbf{Additional Scalability Tests.} For MS-CMR and BraTS, we merged test groups in a similar way to evaluate BInGo (trained via complete learning) on image groups with different sizes. As shown in \cref{fig:large_group_results}, with $M'$ increasing, co-registration becomes significantly more difficult (indicated by the worse initial DSC/gWI), whereas BInGo maintained decent performance for ultra-large groups (e.g., over 1300 images from MS-CMR), and achieved even better accuracy for larger 3D image groups from BraTS, showing remarkable robustness on large-scale groupwise registration. 

\noindent\textbf{Qualitative Results.} We visualized the results from the best baselines (APE/ CTE+AttResUNet/$\mathcal{X}$-CoReg for BraTS/MS-CMR/Learn2Reg) and BInGo in \cref{fig:vis_results}. BInGo could achieve better alignment for both large-scale anatomy and local fine structures in most cases, and the predicted deformations reached great smoothness and diffeomorphism. The feature maps from BInGo were nearly modality-invariant and shared similar structures, illustrating successful disentanglement of the common structure, spatial transformations and appearance information. Besides, the relatively poor performance in certain local areas may be due to the too severe initial misalignment, which made different tissues with similar intensities appear in the same location. Still, BInGo performed no worse than the baselines in such regions, and achieved satisfactory accuracy for the entire foreground. 

\section{Conclusion and Discussion}
In this work we have presented BInGo, a generative Bayesian framework for unsupervised multimodal groupwise registration.
This new formulation of image registration has achieved comparable performance with similarity-based iterative and unsupervised methods.
In particular, we demonstrated that equipped with unique scalability, BInGo could reduce the computational burden of groupwise registration without compromising accuracy. 
This opens up the possibility to realize learning-based multimodal groupwise registration on a large scale and with various group sizes.
A potential limitation of our work is that there may exist performance drop on specific datasets, e.g. the highly challenging abdominal images.
Future work includes investigation into the negative factors that may inhibit BInGo from generalizing to large image groups.

\bibliographystyle{splncs04}
\bibliography{biblio.bib}

\end{document}